\documentclass[conference]{IEEEtran}
\IEEEoverridecommandlockouts
\usepackage{cite}
\usepackage{amsmath,amssymb,amsfonts}
\usepackage{algorithmic}
\usepackage{graphicx}
\usepackage{textcomp}
\usepackage{xcolor}
\usepackage{url}
\def\BibTeX{{\rm B\kern-.05em{\sc i\kern-.025em b}\kern-.08em
    T\kern-.1667em\lower.7ex\hbox{E}\kern-.125emX}}
\begin{document}

\title{ICDM 2020 Knowledge Graph Contest: Consumer Event-Cause Extraction}

\author{
\IEEEauthorblockN{  Congqing He \footnotemark$\dagger$\qquad  }
\IEEEauthorblockA{  \textit{JD Digits} \qquad  \\
  Beijing, China \qquad  \\
  hecongqing@hotmail.com  \qquad  }
\and

\IEEEauthorblockN{\qquad  Jie Zhang \footnotemark$\dagger$ \qquad    }
\IEEEauthorblockA{\qquad  \textit{Alibaba Inc.} \qquad\\
\qquad  Hangzhou, China \qquad \\
\qquad  zhangjie986862436@gmail.com      \qquad   }
\and

\IEEEauthorblockN{Xiangyu Zhu $\ddagger$}
\IEEEauthorblockA{\textit{JD Digits} \\
Beijing, China \\
yuconan@163.com}
\and
\\
\and
\IEEEauthorblockN{ \qquad  \qquad \qquad \qquad \qquad   Huan Liu}
\IEEEauthorblockA{\qquad  \qquad \qquad \qquad \qquad \qquad \textit{JD Digits}  \\
\qquad  \qquad \qquad \qquad \qquad \qquad Beijing, China \\
\qquad  \qquad \qquad \qquad \qquad \qquad liuhuan35@jd.com}
\and
\IEEEauthorblockN{ Yukun Huang}
\IEEEauthorblockA{\textit{Tencent Inc.} \\
Shanghai, China \\
austinhuang@tencent.com}
}

\maketitle
\footnotetext[1]{$\dagger$These authors contributed equally to this work.} 
\footnotetext[2]{$\ddagger$Corresponding authors.} 

\begin{abstract}
Consumer Event-Cause Extraction, the task aimed at extracting  the potential causes behind certain events in the text, has gained much attention
in recent years due to its wide applications. The ICDM 2020 conference sets up an evaluation competition that aims to extract events and the causes of the extracted events with a specified subject (a brand or product). In this task, we mainly focus on how to construct an end-to-end model, and extract multiple event types and event-causes simultaneously. To this end, we introduce a fresh perspective to revisit the relational event-cause extraction task and propose a novel sequence tagging framework, instead of extracting event types and events-causes separately. Experiments show our framework outperforms baseline methods even when its encoder module uses an initialized pre-trained BERT encoder, showing the power of the new tagging framework. In this competition, our team achieved 1st place in the first stage leaderboard,  and 3rd place in the final stage leaderboard.
\end{abstract}


\section{Introduction}

The ICDM 2020 Knowledge Graph Contest is a competition-style event co-located with the leading ICDM conference. This paper describes our solution for the consumer event-cause extraction task, and we won 1st place in the first stage leaderboard and 3rd place in the final stage leaderboard. Extracting causes of consumer events \cite{event-centric knowledge, event causes} are the center of attention for many business scenarios such as content advertising, social listening, etc. Take content advertising as an example. Today’s advertisers are not satisfied with the direct exposure of brands or products, and they prefer to embed the content through product features, and subtly inspire consumers to take the initiative to associate their brands or products with arbitrary consumer events. To this end, explicitly extracting causes of consumer events becomes an important technique to build such a system addressing the advertisers’ needs.

The Consumer Event Cause Extraction (CECE) task aims to extract consumer events and the cause of the event from the text of a given brand or product. Traditional methods use a model structure similar to extract machine reading comprehension (MRC) \cite{mrc}. Most of the related work \cite{Emotion-cause} extracted events type and events-cause separately, without considering the dependencies between them.

In this competition, we introduce a fresh perspective to revisit the relational event-cause extraction task and propose a novel sequence tagging \cite{LSTM-CRF} framework, instead of extracting event types and events-causes separately. Experiments show our framework outperforms baseline methods even when its encoder module uses a randomly initialized BERT \cite{bert} encoder, showing the power of the new tagging framework. In this competition, our team achieved 1st place in the first stage leaderboard,  and 3rd place in the final stage leaderboard.

The rest of the paper is organized as follows: Section \ref{2} describes
our solution which contains the model details. In Section \ref{3} and Section \ref{4} , we show the experiments and results of our model. Finally, we conclude our proposed methods in Section \ref{5}.

\begin{figure*}[htbp]
\centerline{\includegraphics[scale=0.8]{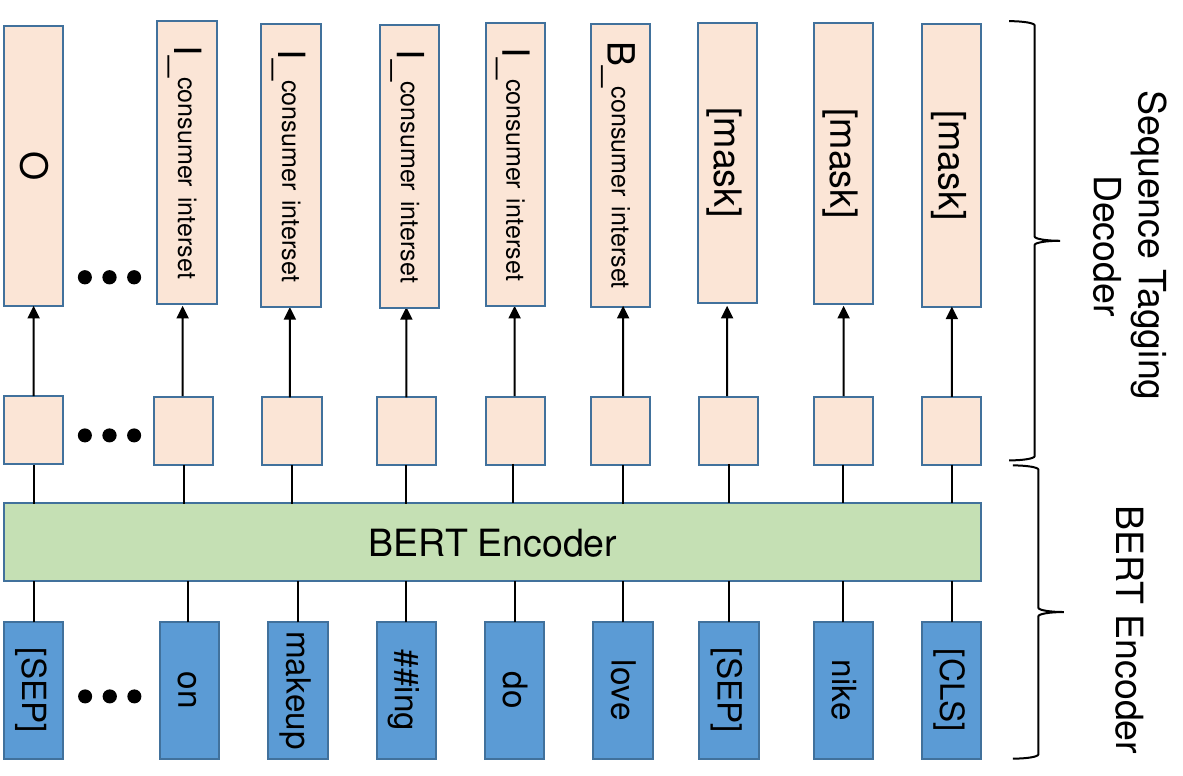}}
\caption{Overview of our model, our proposed  approach for consumer event-cause extraction.}
\label{fig1}
\end{figure*}

\section{Methodology}\label{2}

\subsection{Data Preprocess}

We just removed the ID at the beginning of the content due to the high-quality dataset, such as the \verb|"68771,Love doing makeup on all ages"|, and we converted it to \verb|"Love doing makeup on all ages"|.

\subsection{Model}
To extract consumer event-cause in an end-to-end fashion, Our model is mainly composed of two parts: BERT Encoder and Sequence Tagging Decoder.

\subsubsection{BERT Encoder}
At first, we converted the text and brand/product to  the format of \verb|[CLS] Brand/Product [SEP] Text [SEP]| as inputs $\{x_1,x_2,\cdots,x_n\}$.  At second, we employed a pre-trained BERT model \cite{bert} to encode the content information. The encoder module extracts feature information $z_j$  from sentence $x_j$, then we fed it into the subsequent tagging module.

Here we briefly review BERT, a multi-layer bidirectional Transformer based language representation model. It is designed to learn deep representations by jointly conditioning on both the left and right context of each word, and recently, it has been proven surprisingly effective in many downstream tasks~\cite{BioBERT}~. Specifically, it is composed of a stack of $N$ identical Transformer blocks. We denoted the Transformer block as $Trans(x)$, in which  $X$ represented the input vector. The detailed operations are as follows:

\begin{equation}\label{eq1}
h_0 = SW_s + W_p
\end{equation}
\begin{equation}\label{eq2}
h_l = Trans(h_{l-1}), l \in [1,N]
\end{equation}

where $S$ is th matrix of one-hot vectors of sub-words indices in the input sentence,
$W_s$ is the sub-words embedding matrix, $W_p$ is the positional embedding matrix where $p$ represents the position index in the input sequence, $h_l$ is the hidden state vector, i.e., the context representation of input sentence at $l$-th layer and $N$ is the number of Transformer blocks. Note that in our work the input is a single text sentence instead of sentence pair, hence the segmentation embedding as described in the original BERT paper was not taken into account in Eq. \eqref{eq1}. For a more comprehensive description of the Transformer structure, we refer readers to  \cite{attention}.

\begin{figure*}[h]
\centerline{\includegraphics[scale=0.7]{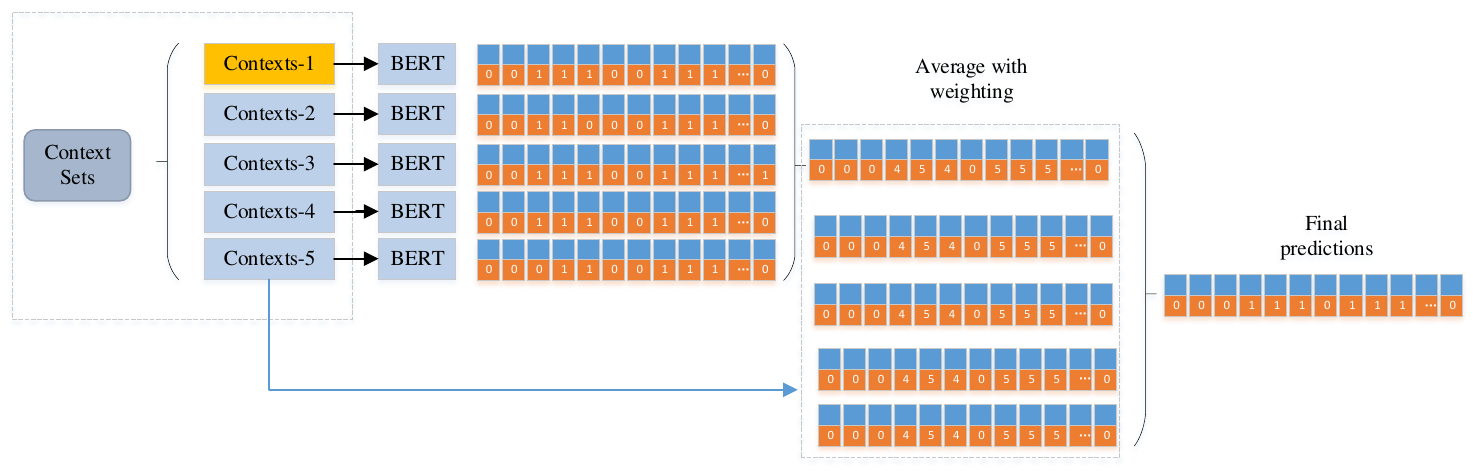}}
\caption{Ensemble strategy based on BERT stacking and weighting operation.}
\label{fig2}
\end{figure*}

\subsubsection{Sequence Tagging Decoder}
In the 2020 ICDM Competition \footnote{\url{https://www.biendata.xyz/competition/icdm_2020_kgc/}}, 
the task adds judgments on multiple event types, which is difficult to solve with a reading comprehension framework. The goal of the competition is to extract multiple event types and event-causes for each text and brand/product. To this end, we proposed a sequence tagging decoder, which can extract multiple event types and event-causes simultaneously.

First, we constructed tags for pairs of input sentences, with a tagger for each token, such as follows:

\noindent\verb|Love doing makeup on all ages, |


\noindent\verb|B_{consumer interest} |

\noindent\verb|I_{consumer interest} ...|

In this way, we can use the softmax function to decode each label independently, and obtain the set of all possible event types and event-causes pairs. Inspired by sequence tagging tasks, it is beneficial to consider the correlations between labels in neighborhoods and jointly decode the best chain of labels for a given input sentence.  Therefore, we modeled label sequence jointly using a conditional random field (CRF) \cite{crf},  instead of decoding each label independently.

Formally, we used $z = \{z_1, z_2, \cdots, z_n\}$ to represent a generic input sequence where $z_i$ is the input vector of the $ith$ word.  $y = \{y_1,y_2,\cdots, y_n\}$ represents a generic sequence of labels for $z$. $\mathbb{Y}(z)$ denotes the set of possible label sequences for $z$. The probabilistic model for sequence CRF defines a family of conditional probability $p(y|z;W,b)$ over all possible label sequences y given $z$ with the following form:
\begin{equation}
    p(y|z;W,b) = \frac{\prod_{i=1}^{n} \psi(y_{i-1},y_i,z)}{\sum_{y^{'} \in \mathbb{Y}(z)} \prod_{i=1}^{n} \psi(y_{i-1}^{'},y_i^{'},z)}
\end{equation}
where $\psi(y^{'},y,z) = exp(W_{y^{'},y}^{T}z_i + b_{y^{'},y} )$  are potential functions , and $W_{y^{'},y}^{T}$ and $b_{y^{'},y}$ are the weight vector and bias corresponding to label pair ${(y^{'},y)}$, respectively.

For CRF training, we used the maximum conditional likelihood estimation. For a training set $\{z_i,y_i\}$, the logarithm of the likelihood (a.k.a. the log-likelihood) is given by:

\begin{equation}
    L(W,b) =  \sum_{i}log(p(y|z;W,b)) 
\end{equation}
Maximum likelihood training chooses parameters so that the log-likelihood $L(W, b)$ is maximized.

Decoding is to search for the label sequence $y^*$ with the highest conditional probability:
\begin{equation}
    y^* = argmax_{y \in \mathbb{Y}(z)}p(y|z;W,b)
\end{equation}

For a sequence CRF model (only interactions between two successive labels are considered), training and decoding can be solved efficiently by adopting the Viterbi \cite{Viterbi} algorithm.



\subsection{Ensemble Methodology}
In the model ensemble \cite{Ensemble} stage, we adopt a simple and efficient way and get $1.30\%$ boosting (Shown in Figure \ref{fig2}). We employed a two-step approach to get the final results. Firstly, we determined the serialization of the text boundary cross-validation results, with the character position of the predicted results being 1 and the rest being 0. Then we stacked all CV results on the corresponding positions, and changed the positions less than $N$ to 0 by threshold.

\section{Experiments}\label{3}

\subsection{Dataset and Metrics}
The ICDM 2020 Knowledge Graph Contest competition provides 500 recent articles\footnote{\url{http://biendata.xyz/media/dataset.zip}} from Instagram which is selected by industry-solution experts to ensure the language formality, diversity, and depth of knowledge in terms of real-world applications.
In addition, the competition focuses on five event types: consumer attention, consumer interest, consumer needs, consumer purchase, and consumer use. The 500 articles will be labeled and served as the training set. There will be a separate online testing set. An example of the data is shown as follows:

\noindent\verb|Input: Text+Brand/Product|
\verb|Output: A span or multiple spans with the|
\verb|predefined event types.|

The competition uses the Precision, Recall, and F1-Measure as the evaluation criteria.
 
\begin{equation}
    precision = \frac{\text{The number of correctly predicted causes}}{\text{The number of predicted causes}} 
\end{equation}
\begin{equation}
    Recall = \frac{\text{The number of correctly predicted causes
}}{\text{The number of gold causes}} 
\end{equation}
\begin{equation}
F1 = \frac{2\times precision \times Recall}{precision + Recall}
\end{equation}

\subsection{Experimental Settings}

We use a  pre-trained  BERT model\footnote{\url{https://storage.googleapis.com/bert_models/2019_05_30/wwm_uncased_L-24_H-1024_A-16.zip}} to initialize the model weight. For training details, We train our model using Adam optimizer with a 2e-5 learning rate and 12 mini-batch sizes, and the L2-norm regularization coefficient is set to 1e-5. We used five-fold cross-validation for each model to improve the model's stability.

Besides, we modeled label sequences jointly using a CRF to improve the performance of our method. The model effect was not improved when the learning rate of CRF was relatively small. To explore the effect of CRF on BERT, we tried to continuously increase the learning rate of CRF.  At last, we concluded that the learning rate of the CRF layer could be improved by increasing the learning rate of the CRF layer.  In the experiment, we set the learning rate of CRF to 1000.

\begin{table*}[h]
\caption{Experimental results of baseline and our model on the ECEC task.}
\begin{center}
\begin{tabular}{|c|c|c|c|}
\hline
\textbf{Dataset}&\textbf{\textit{First Stage Leaderboard}}& \textbf{\textit{Final Stage Leaderboard}} \\
\hline
\textbf{Methods} & \textbf{\textit{F1}}& \textbf{\textit{F1}} \\
\hline
Official Baseline&  0.205 & 0.201 \\
\hline
Our Single Model&  0.325& 0.341 \\
\hline
Our Stacking Model& 0.340&	0.354\\ 
\hline
\end{tabular}
\label{tab1}
\end{center}
\end{table*}

\section{Results}\label{4}

 We firstly introduce the following proposed models:

\begin{itemize}
    \item \textbf{Official Baseline}:This model is based on the pre-trained model BERT and uses a model structure similar to extract machine reading comprehension (MRC).\footnote{The baseline model has been published on github, \url{https://github.com/MAS-KE/ICDM_2020_KGC}}
    \item \textbf{Our Single Model}: This model is based on pre-trained model BERT and uses a novel sequence tagging framework to revisit the event-cause extraction task.
    \item \textbf{Our Stacking Model}: We employ the stacking method to improve model stability and model performance.
\end{itemize}

Main Results Table \ref{tab1} shows the results of different models for CECE on two leaderboards. The single model we proposed, overwhelmingly outperforms the baseline in terms of all leaderboards and achieves encouraging $65.9\%$, $77.0\%$ improvements in F1-score over the baseline method on two leaderboards respectively. Even without taking advantage of the ensemble methodology, our single model is still competitive to the existing baseline. This validates
the utility of the proposed sequence tagging decoder that extracts Consumer Event-Cause.

\section{Conclusion}\label{5}
In this competition, we introduce a fresh perspective to revisit the relational event-cause extraction task and propose a novel sequence tagging framework, instead of extracting event types and events-causes separately. Experiments show our framework outperforms baseline methods even when its encoder module uses a randomly initialized BERT encoder, showing the power of the new tagging framework. In this competition, our team achieved 1st place in the first stage leaderboard,  and 3rd place in the final stage leaderboard.

\section{Acknowledgements}
We thank everyone associated with organizing and sponsoring the 2020 ICDM Knowledge Graph Contest.  Challenge was sponsored and managed by the Mininglamp Academy of Sciences and the Institute of Automation, Chinese Academy of Sciences. The competition platform was hosted by Biendata. We are very grateful to the organizing team Xindong Wu and  Kang Liu for their great efforts during the challenge.

\vspace{12pt}

\end{document}